\documentclass[twoside,11pt]{article}

\usepackage{jmlr2e}
\usepackage[dvipsnames]{xcolor}
\usepackage{amsmath}
\usepackage{graphicx}
\usepackage{caption}
\usepackage{subcaption}
\graphicspath{ {./figs/} }
\usepackage{float}
\usepackage{multirow}
\usepackage{tabularx, ragged2e}
\usepackage{tabu}
\usepackage{enumitem}
\usepackage{hyperref}

\jmlrheading{1}{2019}{1-48}{4/00}{10/00}{John Just}

\ShortHeadings{Deep Generative Models Strike Back!}{Just and Ghosal}
\firstpageno{1}

\begin{document}

\title{Deep Generative Models Strike Back! Improving Understanding and Evaluation in Light of Unmet Expectations for OoD Data}

\author{\name John Just \email justjo@iastate.edu \\
       \addr ISU Digital Ag\\
       Iowa State University\\
       Ames, IA 50010, USA
       \AND
       \name Sambuddha Ghosal \email sghosal@media.mit.edu \\
       \addr MIT Media Lab\\
       Massachusetts Institute of Technology\\
       Cambridge, MA 02139, USA}

% \editor{}
\maketitle

\begin{abstract}%   <- trailing '%' for backward compatibility of .sty file
Continued advances in deep generative and density models have shown impressive capacity to model complex probability density functions (PDF) in lower-dimensional space, and likewise impressive generation of images.  Simultaneously, applying such models to high-dimensional image data with the intent to model the PDF has shown unexpectedly poor generalization, with out-of-distribution (OoD) data being assigned equal or higher likelihood than in-sample data.  Workaround methods to deal with this have been proposed that deviate from a fully unsupervised approach, requiring large ensembles or additional knowledge about the data (such as known outliers) that is not commonly available in the real-world.  In this work, the previously offered reasoning behind these issues is challenged empirically, and it is shown that data-sets such as MNIST fashion/digits, and CIFAR10/SVHN are linearly or trivially separable and thus have no overlap on their respective data manifolds that explains the higher OoD likelihood phenomenon.  Newer deep generative flow models such as  masked autoregressive flows (MAFs) and block neural autoregressive flows (BNAFs) are shown to not suffer from OoD likelihood issues nearly to the extent of previously tested models such as GLOW, PixelCNN++, and real NVP.  Additionally, a new avenue is explored which involves a change of basis to a new space of the same dimension with an orthonormal unitary basis of eigenvectors (e.g. using SVD) prior to modeling.  In the tested data-sets and models, this appears to aid in pushing down the relative likelihood of the contrastive OoD data set and improve discrimination results, in some cases considerably.  In doing so distances are preserved and thus the significance of the density of the original space is maintained, while invertibility remains tractable and trivial for generative models.  Finally, a look to previous generation of generative models in the form of probabilistic principal component analysis (PPCA) is inspired, which is revisited for the same data-sets and shown to work exceptionally well at least for discriminating anomalies based on likelihood in a fully unsupervised fashion compared with pixelCNN++, GLOW, and real NVP with much less complexity and significantly faster training.  Similarly, dimensionality reduction using PCA is shown to improve anomaly detection in deep generative models. 
\end{abstract}

\section{A Not So Long Time Ago....}
The abundance of available data and exponential growth of computational power, combined with the arrival of novel learning methods, has led to a number of breakthroughs in many scientific areas~\citep{esteva2017dermatologist,yamins2016using,alipanahi2015predicting,mnih2015human,silver2016mastering,ghosal2017high,ghosal2017engineering,ghosal2018explainable,ghosal2019weakly,pokuri2019interpretable,shah2019encoding}. The most commonly and widely used learning technique is supervised learning where the learning algorithm utilizes available data to learn from and then perform prediction, classification, segmentation, detection and regression tasks. Supervised learning however, requires these data-sets to be properly curated and annotated, where each data-point (sample) has to have a task-specific label or attribute or annotation associated with it. However, in many applications, and in critical ones, for instance in healthcare, security analysis (to name a few), availability of such labeled data is scarce. This is where unsupervised learning comes into play. One of the many areas where unsupervised learning can prove its mettle in using unsupervised density and generative models for anomaly detection~\citep{liu2016unsupervised, liu2017unsupervised, liu2016unsupervisedB}, clustering, representation learning~\citep{radford2015unsupervised}, and probabilistic (conditional) regression are hardly new concepts. Principal and Independent Component Analysis (PCA/ICA) \citep{ICA:01}, Gaussian Mixture Models (GMMs), and Mixture Density Networks (MDNs)~\citep{bishop:94:mdn} have served as core components and inspirations for autoencoders and variational autoencoders (nonlinear PCA)~\citep{vae:13}, normalizing flows (Nonlinear ICA)~\citep{ICA:01, dinh:2014:nice, papamakarios:2017:maf}, and conditional autoregressive models with neural networks~\citep{germain:19:made},  Use of flexible generative models for OoD detection has many potential uses commercially that include novel ways to minimize the need for annotated data and evaluating the confidence of predictions for machine learning models on new data.  While in theory they can learn highly complex distributions, in practice a phenomenon has been noted in some cases where, unintuitively, OoD data is assigned higher likelihood than the data in which the deep generative model was trained~\citep{choi:19:waic,hendrycks:19:anom,nalisnick:19:anom,shafaei:19}. This has so far been primarily reported with high dimensional image data, although continuing evaluation may yet find a more systemic problem.  In the meantime explanations have been proposed which assign the reason to the OoD data essentially ``sitting inside of” the training data (i.e. having the same mean but lower variance than the training data), thus encouraging an overall higher mean likelihood scores~\citep{nalisnick:19:anom,choi:19:waic}.  However, these explanations appear based on examining the mean and variance of each individual dimension, not the joint distribution.  In three dimensional space, it is not difficult to\textit{ conjure up} counter-examples such as a sphere inside of a toroid that is fully separable yet has the same mean and lower variance in each dimension.  Extending this same concept to a hyper-cube of the dimensions of the images in MNIST is then trivial conceptually, although training models may still suffer from factors noted in the conclusion of~\citep{nalisnick:19:anom, choi:19:waic} such as optimization, model initialization, and architecture (of which there are many more options than those attempted in aforementioned works).  In the case of MNIST vs Fashion MNIST, a human is easily able to tell the difference between either data-set, which suggests that the joint distributions of each data-set are reasonably separable.  Thus, intuitively, images from MNIST should not be assigned higher likelihood values than images from Fashion MNIST if the model was trained on Fashion MNIST.  However, there are ways to quantify the separation that are explored herein.
\noindent

The current methods to evaluate generative models is to compare likelihoods, and observe the density and samples of a trained model on complex 2-dimensional data.  It is already well known that higher likelihood does not necessarily correspond to better sample quality in high-dimensional data~\citep{noteGenEval:16}.  In light of the failures to identify OoD data with adequate confidence, the assumption that achieving a higher likelihood on any given high-dimensional data will translate to better identification of OoD data is explored and challenged here as well. Of course, a good approximation of the true distribution will indeed produce good samples and assign low likelihood for OoD data compared with in-sample data, but the converse statement is not guaranteed.  Furthermore, a model considered satisfactory for most real-world use cases will (among other things) need to provide support for anomaly detection, novel data discovery, assigning confidence of model predictions to new data, and use as Bayesian proposal distributions.  Thus the aim of the research presented here is not to necessarily obtain higher state of the art (SOTA) likelihood values on popular tabular or image data-sets.  Although some of the methods presented here might be useful for that, simply achieving higher likelihood does not help put practitioners in a better position to have confidence that employing these unsupervised learning models for the aforementioned real-world tasks can be done with a high level of confidence in the absence of adversarial data to test against, especially for high-dimensional data.  Instead, a combination of methods are presented that can be utilized to support the aforementioned objectives, and also provide some degree of diagnostics and intuition, all within a full unsupervised framework.

\subsection{Background}
\subsubsection{Normalizing Flows}
The models used in this investigation and~\citep{nalisnick:19:anom} are considered as normalizing flow-type models. ``Normalizing" refers to the transformation of the PDF of the data following the application of the ``change of variables" technique. ``Flows" alludes to the ability to compose (or stack) the transformations, leading to more complex transforms.  The change of variables theorem states that we may use a bijective function $f_{\theta}: \textbf{X}\to\textbf{Y}$, where $\textbf{X}, \textbf{Y}\subseteq \mathbb{R}^{d}$ are two different probability distributions and $\theta$ parameterizes the function $f$, to transform a continuous random variable (R.V.) $\textbf{x}\sim\textbf{X} \Rightarrow\textbf{y}\sim\textbf{Y}$ such that  the PDFs of the two R.V.s are related by $p_{Y}(y)|det\textbf{J}_{f_{\theta}(x)}|=p_{X}(x)$.  To remain tractable, the methods studied here restrict dependencies among the variables within the function $f$ to ensure the Jacobian is lower triangular (thus the determinant is simply the trace of the Jacobian).  This is satisfied by a conditional, or autoregressive, dependency structure in which the $i^{th}$ output of the transform $y_{i}=f_{\theta}^{(i)}(x_{\leq i})$ with $i\leq d$.  Note that any functional dependence of $y_{i}$ on $x_{i}$ must additionally be monotonic to satisfy the bijective requirement.  Real NVP and MAF simply use $x_{<i}$ as inputs to $f_{\theta}^{(i)}$, which is a neural network that outputs a scale and shift value for $x_{i}$ for each flow.  BNAF also parameterizes a neural network for $f_{\theta}^{(i)}$ and includes $x_{i}$ as an input, but forces the weights that interact directly with it to be positive such that monotonicity is guaranteed.

\subsubsection{Connecting the Different Generative Models}
In terms of the relationships between the various generative models utilized and referenced in this and cited works, Real NVP~\citep{dinh:17:realNVP} is simply a restricted form of MAF~\citep{papamakarios:2017:maf} in which some of the dependencies have been left out.  GLOW \citep{glow:2018} is equivalent to real NVP except for the permutations of the R.V.s between flows is replaced with an invertible linear transformation.  MAF is just a stack (composition) of MADEs~\citep{germain:19:made} where the R.V.s are permuted between successive flows.  Note that MADE (and PixelCNN++~\citep{Salimans2017PixeCNN}) decompose the joint density as a product of conditional densities per the chain rule, but since they do not permute or mix up the dimensions like the normalizing flows do, they do not require a re-scaling of the probabilities of the outputs by the absolute value of the Jacobian determinant.  BNAF~\citep{bnaf19} is a normalizing flow with similar theoretical underpinnings to the aforementioned models, but the structure of the transformer is such that BNAF is a true universal approximator, which is a desirable property shared by unconstrained neural networks that means it can be made arbitrarily flexible (though this may not be feasible in practice).

\subsection{Contributions}
In this work, the reasons for deep generative models failures to assign a higher likelihood to in-sample data than OoD data is investigated in greater depth for MNIST fashion VS digits, and CIFAR10 vs SVHN data-sets.  The following contributions are offered by this work:
\begin{enumerate}[nolistsep]
\item It is proven that the Fashion MNIST/Digits MNIST data-sets are linearly separable, and the CIFAR10/SVHN data-sets are trivially separable.  These results diverge from previous work indicating that the higher OoD likelihood problem was one in which the MNIST digits and SVHN data-sets were contained within the Fashion and CIFAR10 data-sets, respectively \citep{nalisnick:19:anom}.
\item The higher likelihood of OoD data is shown to be isolated to previous models that used convolutional neural networks and coupling layers (PixelCNN, real NVP, GLOW).  Slightly more recent normalizing flows like MAF and BNAF, which are more flexible and utilize dense-neural networks as transformers in the generative models, do not exhibit the OoD problem as severely (although are shown to still exhibit higher-than-expected OoD Likelihood under certain conditions).
\item Using a baseline of a full-covariance Normal distribution is shown to be highly competitive with modern generative models, and a sorely overlooked option.
\item Given previous literature that has shown that likelihood is uncorrelated with sample quality~\citep{noteGenEval:16}, it is also shown that likelihood is uncorrelated with OoD detection potential.
\item Higher than expected OoD likelihood is confirmed with extensive support to be primarily an optimization problem given a sufficiently flexible density model architecture.  Upon examining several models, we find that changes such as a new basis for the input data can lead to the input distributions becoming more Normal.  Furthermore, reducing dimensionality can offset the effects of the curse of dimensionality.  Both these techniques can lead to significantly better OoD detection performance.  Further intuition during the optimization process is presented, with expected ideal behavior of OoD log-likelihood (LL) during the training process, hypothesized and then shown to be obtainable under certain conditions.
\item Several diagnostics methods such as observing the granularity of the estimated LL and comparing data LL with sample LL are shown to underscore the improvements proposed in this work extend to re-world benefits while maintaining a fully unsupervised approach to employing deep generative models as anomaly and novel data detectors.
\end{enumerate}
%\newpage
\section{The Phantom Menace:  Unexpectedly High OoD Likelihood}
\noindent\subsection{Diagnosing the Problem}
\begin{figure}[H]
  \begin{subfigure}[b]{0.49\textwidth}
    \includegraphics[width=\textwidth]{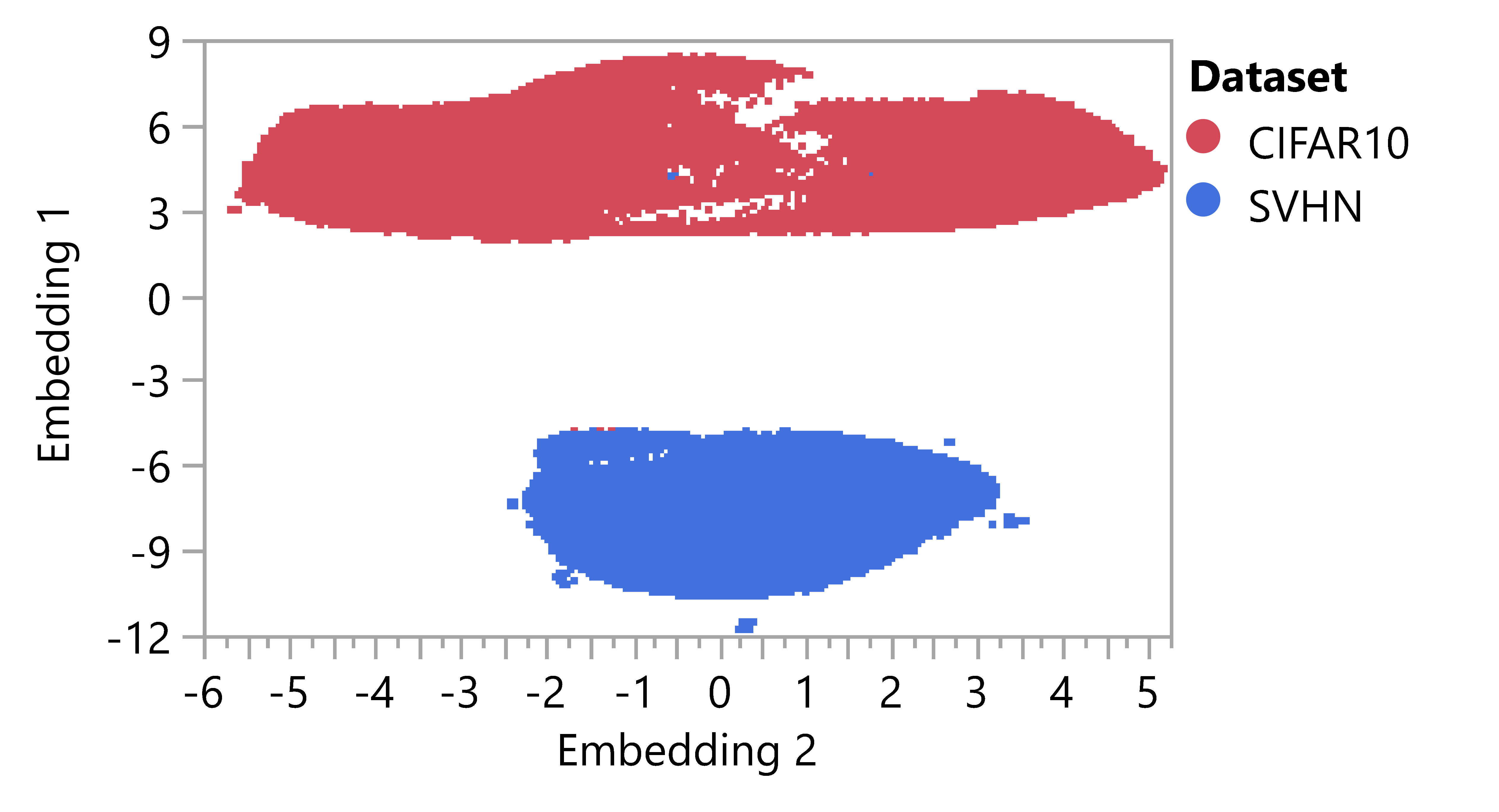}
    \caption{CIFAR vs SVHN.}
    \label{fig:f1}
  \end{subfigure}
  \hfill
  \begin{subfigure}[b]{0.5\textwidth}
    \includegraphics[width=\textwidth]{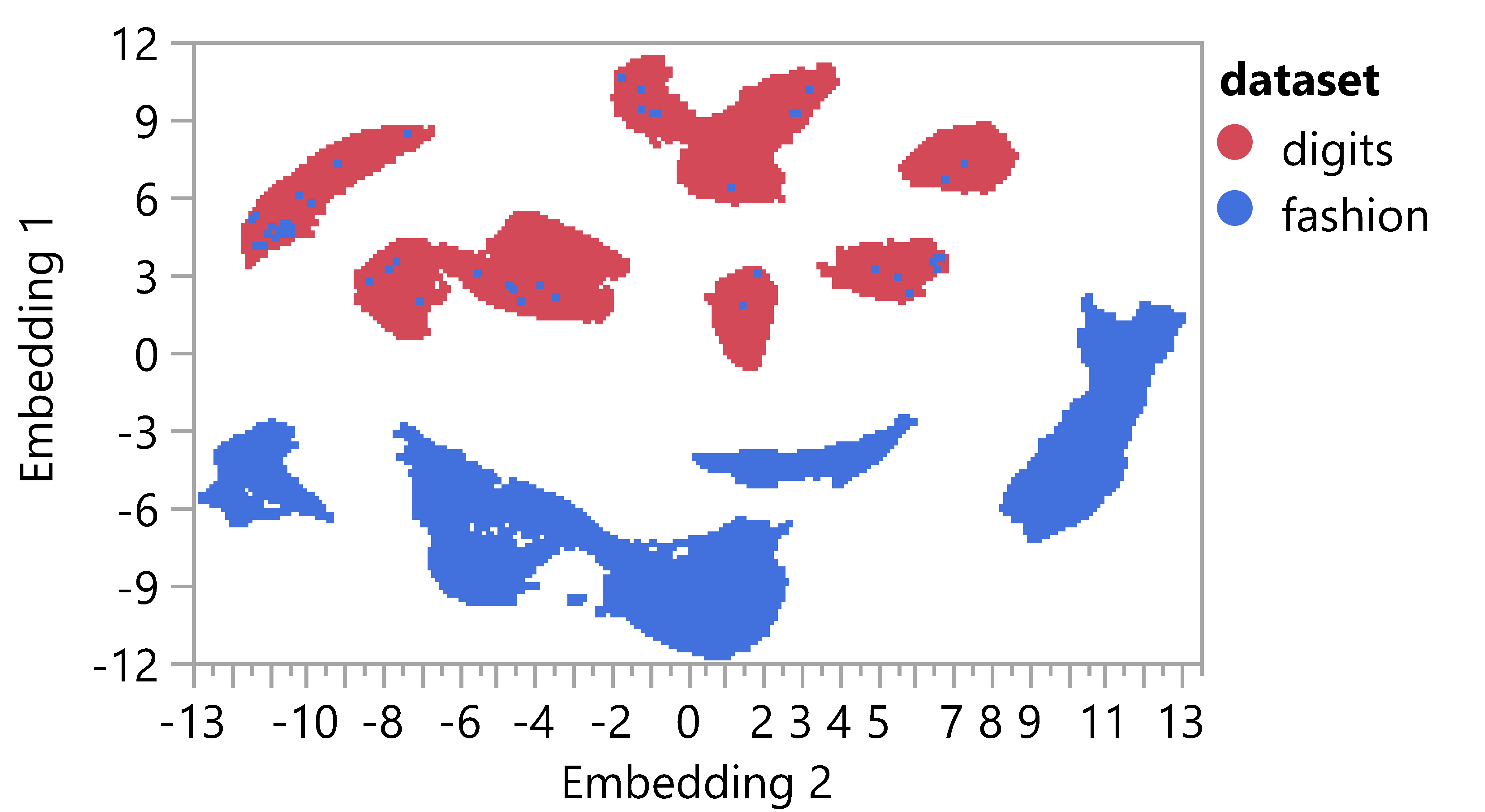}
    \caption{fashion VS digits MNIST.}
    \label{fig:f2}
  \end{subfigure}
  \caption{\label{fig:UMAP} UMAP with default settings is used to find 2d embeddings for each data-set.  The embeddings divide the data-sets almost entirely.}
\end{figure}

\noindent
The primary barrier to identifying the underlying issues observed with density modeling of images has to do with the number of dimensions of the problem.  CIFAR10, which by comparison with most real-world applications consists of relatively small 32x32x3 color images, contains 3072 pixels/dimensions per image.  Fashion MNIST is far smaller yet at 784 pixels, which is still very large.  Visualizing all the raw data at once in more than 2-dimensions is unwieldy at best as dimensions increase, and thus resorting to some form of dimensionality reduction is usually leveraged to explore and observe the relationships.  Projecting the data onto a lower-dimensional manifold by techniques such as PCA are well understood but comes at the cost of information loss and making assumptions about what features of the data are most important.  Projection techniques such as UMAP \citep{2018arXivUMAP} involve assumptions about the smoothness and closeness of the data to set hyperparameters, and then relies on an optimization procedure which can potentially and significantly distort the data density and distances.  Applying UMAP on the CIFAR10/SVHN and fashion/digits MNIST data-sets with default hyperparameter settings produces visually pleasing results (Figure \ref{fig:UMAP}) and divides the data-sets almost entirely.

\noindent Given the reasonable results obtained so easily with UMAP, it does not seem logical that highly flexible generative models should fail to assign a low likelihood to the OoD data-set, but UMAP provides no consistent basis with which to define the level or ease of separability of the data. 

\subsection{Separability}

While exploring various clustering methods may give some intuition about data the similarities and differences of the high-dimensional data, we seek a more rigorous definition of the data separability.  For one definition of linear separability, if we have two sets \(\textbf{A} =\{a_{1},..,a_{N_{1}}\} \subset \mathbb{R}^{M}\) where \(\textbf{A} \in \mathbb{R}^{N_{1}\times M}\) and \(\textbf{B} =\{b_{1},..,b_{N_{2}}\} \subset \mathbb{R}^{M}\) where \(\textbf{B} \in \mathbb{R}^{N_{2}\times M}\), then \(\textbf{A}\) and \(\textbf{B}\) are said to be linearly separable if a hyper-plane \(h \in \mathbb{R}^M\) and scalar \(\beta \in \mathbb{R}\) exist such that \(\textbf{A}h>\beta\)  for all \(a \in \textbf{A}\) and  \(\textbf{B}h<\beta\)  for all \(b \in \textbf{B}\).  This can be formulated into a linear programming (LP) problem with a system of “less than” inequalities as shown in Equation \ref{eq1}, which can efficiently verify upon convergence (if a solution is found) that the data are separable.  Note that an additional buffer/boundary of \(\epsilon\) can be added for visibility purposes since the problem involves a finite set on \(\mathbb{R}^M\). 
\begin{equation} \label{eq1}
\begin{split}
-\textbf{A}h+\beta<-\epsilon  \\
\textbf{B}h - \beta<-\epsilon \\
f=\bar{\textbf{0}}
\end{split}
\end{equation}
\noindent Note that the objective function \(f\), which constrains the solution, is simply set to the 0 vector since we only care about finding any separating plane to prove separability, not something like the maximum margin hyperplane.  In practice, LP can require a large amount of memory though, so an alternative means involves optimizing a linear support vector machine to find a plane that separates the data.  With this we exchange a longer convergence time for lower memory requirements.  Using this method it was confirmed that the MNIST/F-MNIST data-sets are separable.  On the other hand, CIFAR10/SVHN did not completely separate with a linear plane, with about 18\% and 15\% respectively being mis-classified.  Regardless though, the results of using UMAP seemed to support that the data-sets are non-overlapping, even if not linearly separable.  A very simple nonlinear classifier which involves an ANN with one hidden layer and three tanh activation units also indicates the SVHN and CIFAR10 data-sets are trivially separable, achieving 99.1\% and 95.6\% classification accuracy respectively, using only the first 30 (of 3072) principal components.  Thus there is no evidence that the problems from high OoD likelihood stem from overlapping data manifolds.  Instead we focus attention on areas related to optimization and the potential impacts of the curse of dimensionality.

\subsection{Baseline Density Models}
Given the aforementioned evidence that the CIFAR10/SVHN and fashion/digits MNIST data-sets are trivially separable and models such as real NVP, GLOW, and PixelCNN++ failed to behave as expected with OoD data, we wish to evaluate likelihood using a few of the recent generation of density models and observe if the situation is any improved. Unlike typical generative model papers though, improvement is not judged by the magnitude of the LL achieved, but instead on the discrimination between the in-sample test data and the OoD data via a simple linear logistic regression using only likelihood as evaluated by the trained models.  This amounts to an AUC score shown in Table \ref{BASEmnistTable} and Table \ref{BASEcifarTable}, and since the LL is always equal or higher for the test data than OoD data this correlates to anomaly detection potential.  Models are trained on the fashion MNIST data-set in Figure \ref{BASEmnistTable} and on the CIFAR10 data-set in Figure \ref{BASEcifarTable}.  Values in parenthesis (when given) are quantized values, otherwise the pixel data is de-quantized as done in \citep{dinh:17:realNVP}.  The columns of ``Test", ``OoD", and ``Samp" are the mean log-likelihood (LL) values as evaluated on the fashion MNIST (or CIFAR10) test set, digits MNIST (or SVHN) data-set, and samples generated from the models where possible, respectively.  BNAF is not easily sampled from since the inverse is not analytic, and as such, no corresponding likelihood scores for samples are given.  In all cases the desired outcome is higher LL for the test set, lower LL for the OoD data, LL for the samples to be equal to the test set, and AUC score as close to 1.0 as possible.
\vspace{5pt}
\subsubsection{Model Details} 
For all deep generative models, the training data is separated into train/validate, with validation used for early stopping. ADAM~\citep{ADAM15} optimization is employed. The $MAF_5$ model consists of five flows with one hidden layer of 100 relu units per flow, and batch normalization. The $MAF_{10}$ model consists of 10 flows with two hidden layers of 1024 relu units per flow, and batch normalization. BNAF was only used on MNIST and consisted of six flows with one hidden layer and 12 tanh units per block per flow.  BNAF employed weight normalization.  The Gaussian model (and later PPCA) were fit with the defaults of the scikit-learn package in Python.  Links to code repositories to train with BNAF and MAF are as follows:  \href{https://github.com/johnpjust/BNAF_tensorflow_eager/tree/GQC_Images}{\textcolor{blue}{BNAF}}, \href{https://github.com/johnpjust/maf_tf}{\textcolor{blue}{MAF}}.
\vspace{5pt}
\subsubsection{Baseline Results}
\newcolumntype{Y}{>{\centering\arraybackslash}X}
\noindent
\begin{table}[h!]
\caption{Fashion MNIST baseline density modeling results (values in parentheses denote results for models trained on non-quantized values).}
\begin{tabularx}{\textwidth}{@{}lYYYYY@{}}
\hline
  \textbf{Model} & \textbf{Test} &  \textbf{OoD}  & Samp & AUC & \#Params\\
\hline
Normal & 294.4 & -30.7 & 309.3 & 0.865 & 307.7k\\
$MAF_5$&613.3 (737.9)&-516.9(-870.9)&646.8 (747.1)&0.918 (0.931)&1.1M\\
$MAF_{10}$&1749 (2141)&1267 (1191)&1978 (2581)&0.735 (0.775)&34.6M\\
BNAF&1906 (2041)&1943 (2037)&N/A&0.5 (0.52)&44.4M\\
\hline
\end{tabularx}
\label{BASEmnistTable}
\end{table}
\noindent
\begin{table}[h!]
\caption{CIFAR10 baseline density modeling results (values in parentheses denote results for models trained on non-quantized values).}
\begin{tabularx}{\textwidth}{@{}lYYYYY@{}}
\hline
  \textbf{Model} & \textbf{Test} &  \textbf{OoD}  & Samp & AUC & \#Params\\
\hline
Normal&5060.4&-199.3k&5270&1.0&4.7M\\
$MAF_5$&2088 (2089)&-7240 (-7429)&2161 (2125)&1.0 (1.0)&4.6M\\
$MAF_{10}$&4776 (4795)&-1471 (-1341)&5018 (4999)&1.0 (1.0)&105M\\
\hline
\end{tabularx}
\label{BASEcifarTable}
\end{table}
\\
\noindent
\newpage
There are several interesting findings regarding the baseline models:
%\vspace{-20mm}
\begin{itemize}
\item In contrast with those tested in \citep{hendrycks:19:anom,nalisnick:19:anom,choi:19:waic,shafaei:19}, all models tested here obtain mean in-sample (test) log-likelihood (LL) scores at least at high or higher than the OoD data-set.  
\item CIFAR10 is easily distinguished from SVHN with MAF or simply using a full-covariance Normal distribution.  BNAF requires too many parameters to train on CIFAR10, but when tested on MNIST it performs poorly relative to the MAF and full-covariance Normal models.
\item There is clearly no relation between the mean LL and the ability of the learned model to discriminate outliers, at least as judged by the chosen OoD data-set.
\item While increasing model capacity/flexibility leads to higher likelihoods, it does not directly translate to better discrimination of anomalous data.  In all cases except $MAF_5$ vs full-covariance Normal for MNIST, increasing model capacity is inversely correlated with anomaly detection potential as judged by the AUC scores and difference between Test and OoD likelihood scores.
\item The parsimonious full-covariance Gaussian model is extraordinarily good as an unsupervised discrimination model relative to its complexity, achieving very competitive AUC scores for the data-sets tested in this work.  When compared with the training complexity of the other models in the table, it stands out even further.  Furthermore, the test LL on CIFAR10 is also competitive.  It is quite surprising that this option was never included as a baseline in any of the previous deep generative modeling literatures cited by this work.
\end{itemize}
\vspace{-10mm}
\subsection{Diagnostics and Setting Expectations}
\begin{figure}[H]
  \begin{subfigure}[b]{0.49\textwidth}
    \includegraphics[width=\textwidth]{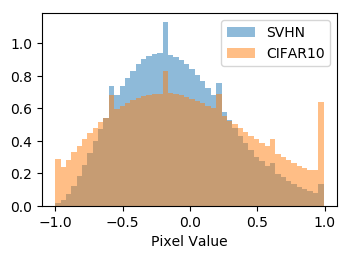}
    \caption{CIFAR vs SVHN.}
    \label{fig:f3}
  \end{subfigure}
  \hfill
  \begin{subfigure}[b]{0.5\textwidth}
    \includegraphics[width=\textwidth]{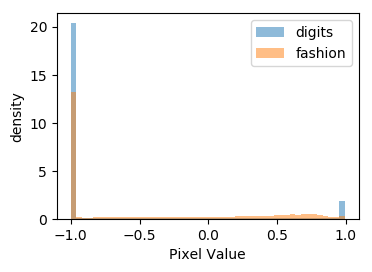}
    \caption{MNIST fashion vs digits.}
    \label{fig:f4}
  \end{subfigure}
    \caption{\label{fig:pixeldist} Pixel distributions for each data-set used to train models (CIFAR10, fashion MNIST) and the corresponding OoD data-set overlaid.  Values on the x-axis correspond to the original pixel value/128 - 1.  Note that the y-axis in each plot is normalized such that the histograms are a valid pdf.}
  \label{fig:trainingIterOrig}
\end{figure}

%\vspace{-10mm}
The findings from the baseline modeling suggest further investigation into the optimization process, and potentially making the data/network more amenable to learning.  Substantial gains in optimizing neural networks have generally been made when techniques to make the data more Normally distributed are employed~\citep{glorot:10, batchnorm:15}. The distributions of the image data-sets shown in Figure \ref{fig:pixeldist} deviate far from Normality, with MNIST (Figure \ref{fig:f2}) far more so than CIFAR10/SVHN (Figure~\ref{fig:f1}). This is interesting since the MNIST data-sets were substantially harder for all the networks to learn (Table~\ref{BASEmnistTable}) to discriminate OoD data compared with CIFAR10/SVHN (Table~\ref{BASEcifarTable}),  but yet are linearly separable data-sets while CIFAR10/SVHN are not.

\begin{figure}[H]
  \begin{subfigure}[b]{0.49\textwidth}
    \includegraphics[width=\textwidth]{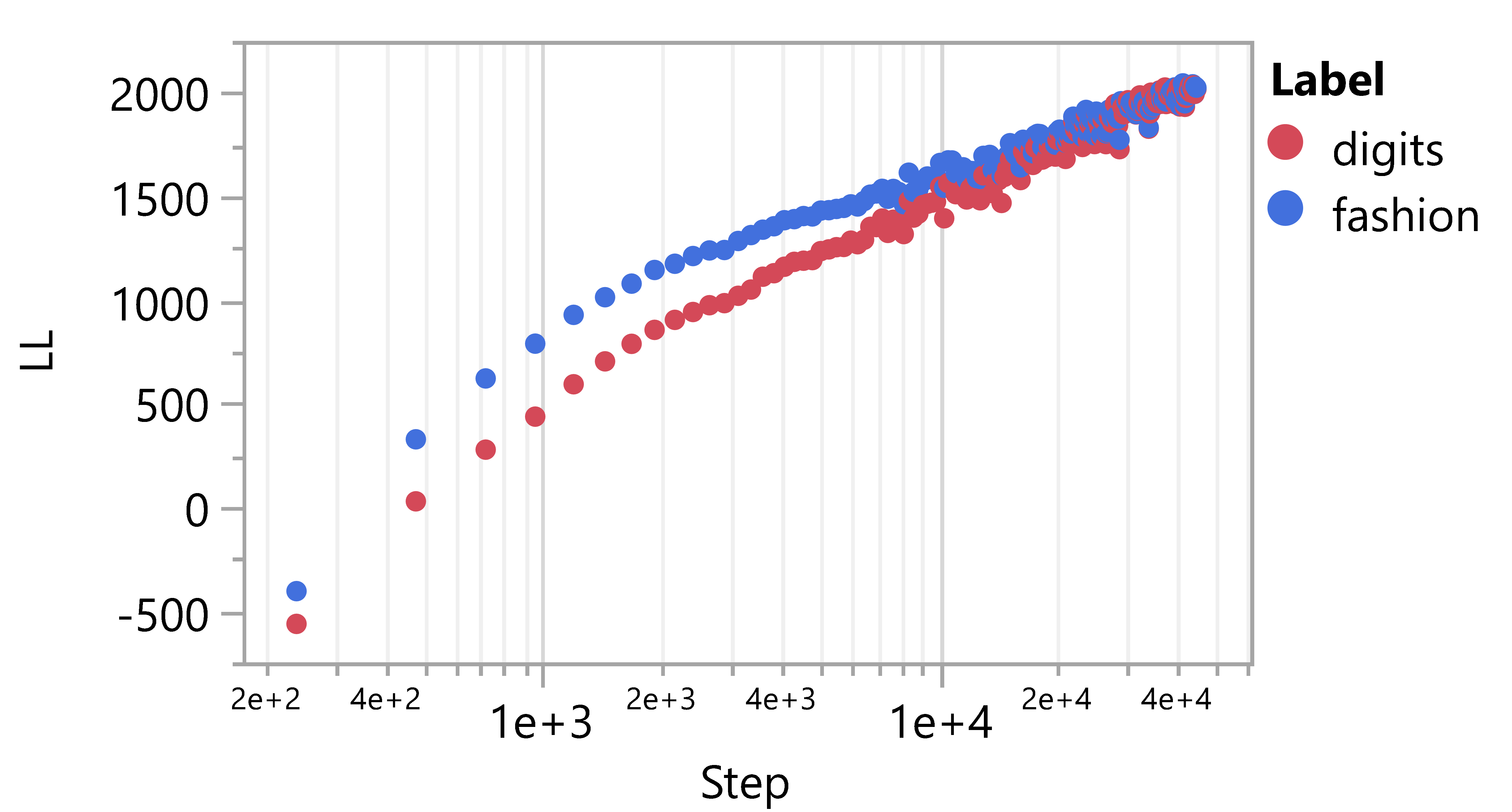}
    \caption{Train=fashionMNIST, OoD=digits}
    \label{fig:BNAFtrainstepsFMNIST}
  \end{subfigure}
  \hfill
  \begin{subfigure}[b]{0.5\textwidth}
    \includegraphics[width=\textwidth]{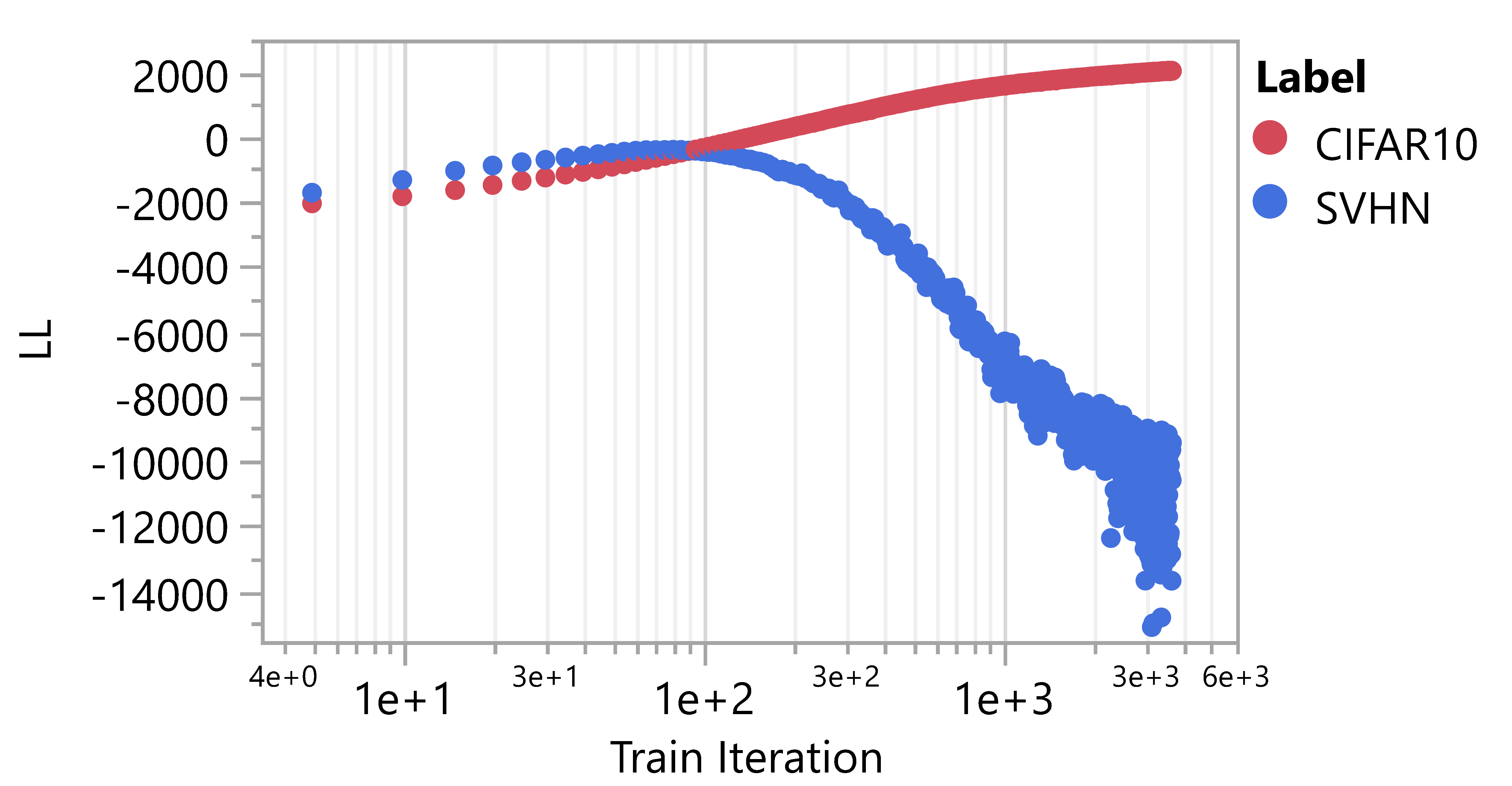}
    \caption{Train=CIFAR10, OoD=SVHN}
    \label{fig:MAF5trainstepsCifar}
  \end{subfigure}
    \caption{\label{fig:baselineTraining} The LL of the validation and OoD data-sets throughout the training process is shown for BNAF on fashion MNIST in Figure \ref{fig:BNAFtrainstepsFMNIST} and $MAF_5$ on CIFAR10 in Figure \ref{fig:MAF5trainstepsCifar}}
  \label{fig:MODELtrainSteps}
\end{figure}

\begin{figure}[H]
  \begin{subfigure}[b]{0.49\textwidth}
    \includegraphics[width=\textwidth]{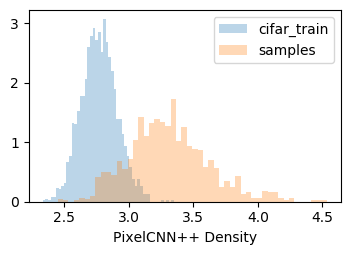}
    \caption{PixelCNN++ Samples, Train Data Density}
    \label{fig:pixelcnnDensityCifar}
  \end{subfigure}
  \hfill
  \begin{subfigure}[b]{0.5\textwidth}
    \includegraphics[width=\textwidth]{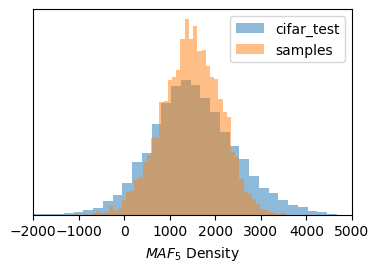}
    \caption{MNIST fashion vs digits.}
    \label{fig:MAF5densityCifar}
  \end{subfigure}
    \caption{Distributions of the likelihood (pixelcnn++ likelihood scaled for visual convenience) for CIFAR10 is shown for the train/test images and the samples from the trained model.  The large disparity between training and sampled data likelihood distributions for pixelcnn++ indicates a poor model fit, whereas the MAF model indicates significantly more confidence in the model.}
  \label{fig:origDataDist}
\end{figure}

\noindent 
In training the density estimation models, a reasonable expectation would be that both in-sample and OoD data start with roughly the same [low] LL before beginning training.  Since both the in-sample and OoD data-sets are images and more similar than random data over the support in $\mathbb{R}^{M}$ (where $M$ is the dimensionality of the data), the LL should initially increase for both data-sets as training commences and then decreases rapidly for the OoD data as the modeled PDF contracts likelihood around the training data manifold.  This is in fact what happens for the CIFAR10 density model in Figure \ref{fig:MAF5trainstepsCifar} using MAF, but not for the  BNAF model on fashion MNIST in Figure \ref{fig:BNAFtrainstepsFMNIST}.
\vspace{-8pt}
%\noindent
\vspace{-4mm}
A necessary, but not sufficient, condition for the learned PDF to represent the actual unknown PDF well is that the likelihood of samples drawn from the model and evaluated by it has a histogram which matches the training data.  Normalizing flow models enable limited diagnostics in low dimensions, such as scatter plots of the data after transforming to the simple distribution.  However viewing the likelihood histograms can help even in the case of autoregressive models or where it is difficult to sample from a model, like BNAF.  For instance, PixelCNN++ has been noted to be a poor anomaly detector\citep{shafaei:19}, and we find that samples from the model achieve much higher LL scores than the training data Figure \ref{fig:pixelcnnDensityCifar}.  In contrast, when trained with $MAF_5$ we find the LL scores of the samples much more closely match that of the data in Figure \ref{fig:MAF5densityCifar}.  It is with these expectations (and noted deficiencies) that we search for ways to improve the results.

\section{A New Hope:  A New Basis}
\begin{figure}[H]
  \begin{subfigure}[b]{0.49\textwidth}
    \includegraphics[width=\textwidth]{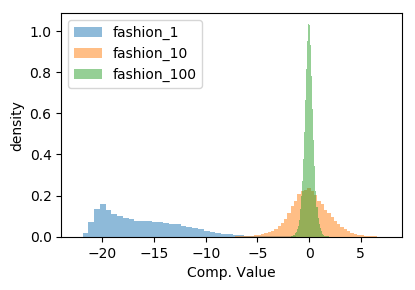}
    \caption{Fashion MNIST.}
    \label{svdpixeldistplots:1}
  \end{subfigure}
  \hfill
  \begin{subfigure}[b]{0.5\textwidth}
    \includegraphics[width=\textwidth]{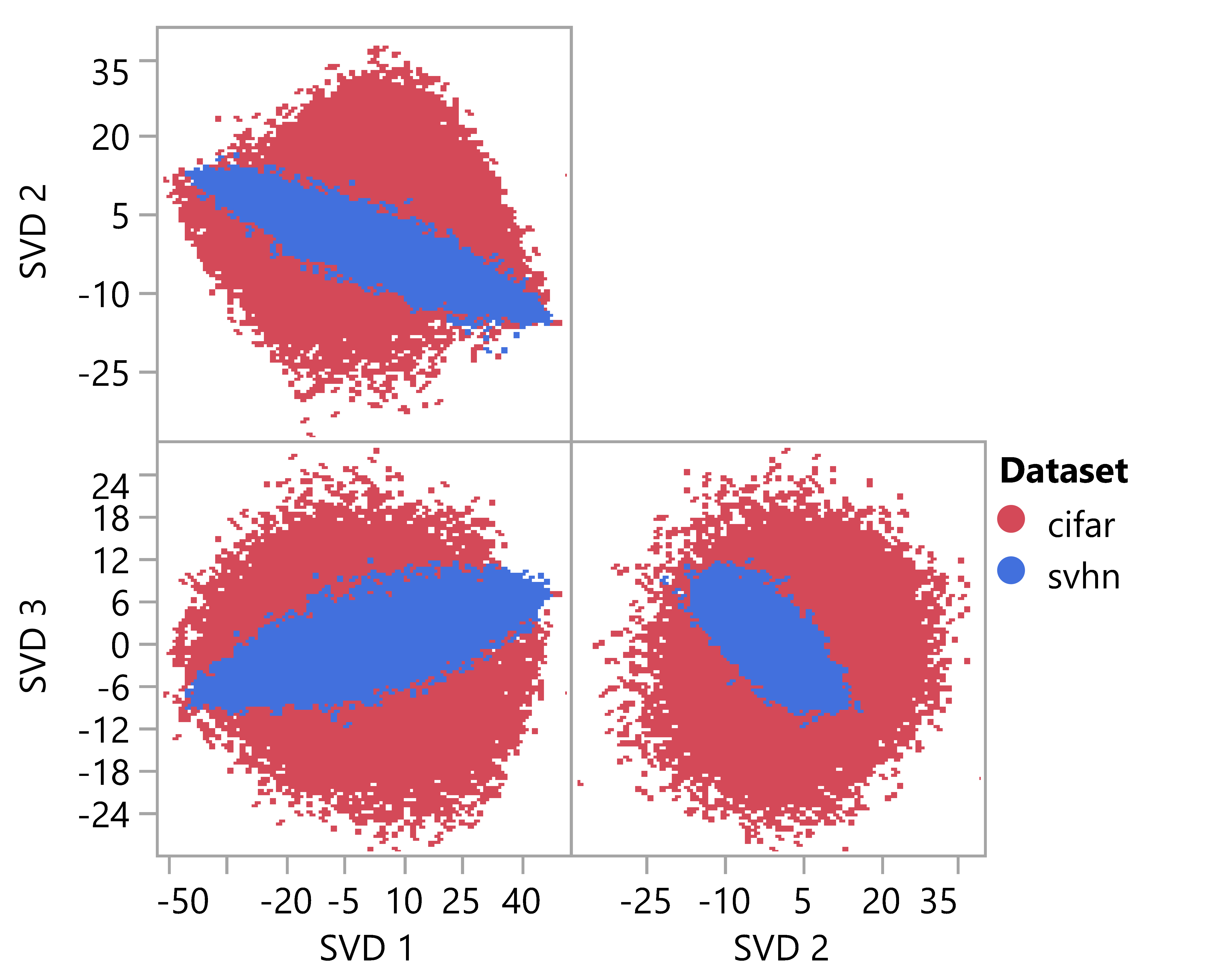}
    \caption{CIFAR10 and SVHN}
    \label{svdpixeldistplots:2}
  \end{subfigure}
  \caption{Feature distributions after using the orthonormal basis found via factorizing the data with SVD.  \ref{svdpixeldistplots:1}) MNIST data-set values on the x-axis correspond to the feature values (after changing basis).  The y-axis in each plot is normalized such that the histograms are a valid pdf.  Several dimensions are overlaid. \ref{svdpixeldistplots:2}) A scatter plot showing the first few dimensions of CIFAR10 overlaid with SVHN, again after re-basing the pixel data.}
  \label{fig:svdDist}
\end{figure}

There is no doubt that the increasing theoretical capacity of the models is linked to better representation of the unknown density function in lower-dimensional space, as is shown in each incremental improvement in literature.  However, we also know that as models grow in size they are more difficult to train, and the curse of dimensionality compounds the issue.  Batch normalization \citep{batchnorm:15}, network initialization \citep{glorot:10}, and data normalization have been used extensively and successfully to combat these issues.  The goal of those procedures is to make the data and network architecture more amenable to learning, and the more Normal (Gaussian) data is typically the easier it is for the networks to learn (optimize).  For some data, pre-processing with nonlinear transforms can help immensely, but this changes the relative density information that we seek to capture with generative models.  Thus, we desire to make the data distributions appear more Gaussian to ease the learning process, but also want to preserve the relative distances of the data points in Euclidean space such that the unknown PDF of the data remains unchanged.  This is possible if we re-basis the data using an orthonormal basis.  Singular value decompositions (SVD) is one such option to do this.  SVD factors an $m \times n$ matrix $\textbf{M}=\textbf{U} \Sigma \textbf{V}$, where $\textbf{U}$ and $\textbf{V}$ are $m \times m$ and $n \times n$ unitary orthonormal matrices, respectively, and $\Sigma$ is a diagonal matrix comprised of the square roots of the eigenvalues of the covariance matrix of $\textbf{M}$.  In our case, $\textbf{V}$ is what we want, since it is a set of orthonormal vectors of the same dimensionality as the data, that are also conveniently ordered by their eigenvalues.  It is also unitary and thus trivial to invert so that images can be constructed from samples.  The results of using this to re-basis the images from this work can be seen in Figure \ref{fig:svdDist} and compared with Figure\ref{fig:origDataDist}.

\noindent
Observing pre and post application of the change of basis, it is clear there has been a significant change in the pixel distributions such that they look far more Normally distributed.  Overall this had a very significant and positive impact on the AUC scores, with mixed results on the LL scores, as seen in \ref{SVDmnistTable} and \ref{SVDcifarTable}.  In all cases except for the $MAF_5$ model the AUC scores increased (or in the case of CIFAR10 vs SVD, the separation between Test and OoD likelihood increased, since AUC was already ~1.0).  This continues to underscore that there isn't a correlation between LL and anomaly detection potential with generative models.  
\noindent
\begin{table}[h!]
\caption{PDF modeling results of fashion VS digits MNIST after re-basing the data using an orthonormal matrix found by SVD (values in parentheses denote results for models trained on non-quantized values).}
\begin{tabularx}{\textwidth}{@{}lYYYYY@{}}
\hline
  \textbf{Model} & \textbf{Test} &  \textbf{OoD}  & Samp & AUC & \#Params\\
\hline
$MAF_5$&552.9 (445.7)&141.9 (85.5)&529.8 (461.9)&0.843 (0.851)&1.1M\\
$MAF_{10}$&923.2 (958.8)&308.0 (331.1)&976.7 (1057)&0.862 (0.859)&34.6M\\
BNAF&1406 (1563)&674.0 (827.4)&N/A&0.83 (0.80)&44.4M\\
\hline
\end{tabularx}
\label{SVDmnistTable}
\end{table}
\noindent
\begin{table}[h!]
\caption{PDF modeling results of CIFAR10 VS SVHN after re-basing the data using an orthonormal matrix found by SVD (values in parentheses denote results for models trained on non-quantized values).}
\begin{tabularx}{\textwidth}{@{}lYYYYY@{}}
\hline
  \textbf{Model} & \textbf{Test} &  \textbf{OoD}  & Samp & AUC & \#Params\\
\hline
$MAF_5$&5668 (5678)&-inf (-inf)&5797 (5813)&1.0 (1.0)&4.6M\\
$MAF_{10}$&5716 (5759)&-inf (-inf)&5902 (5860)&1.0 (1.0)&105M\\
\hline
\end{tabularx}
\label{SVDcifarTable}
\end{table}
\\
\section{Return of PPCA}
\begin{figure}[H]
  \begin{subfigure}[b]{0.58\textwidth}
    \includegraphics[width=\textwidth]{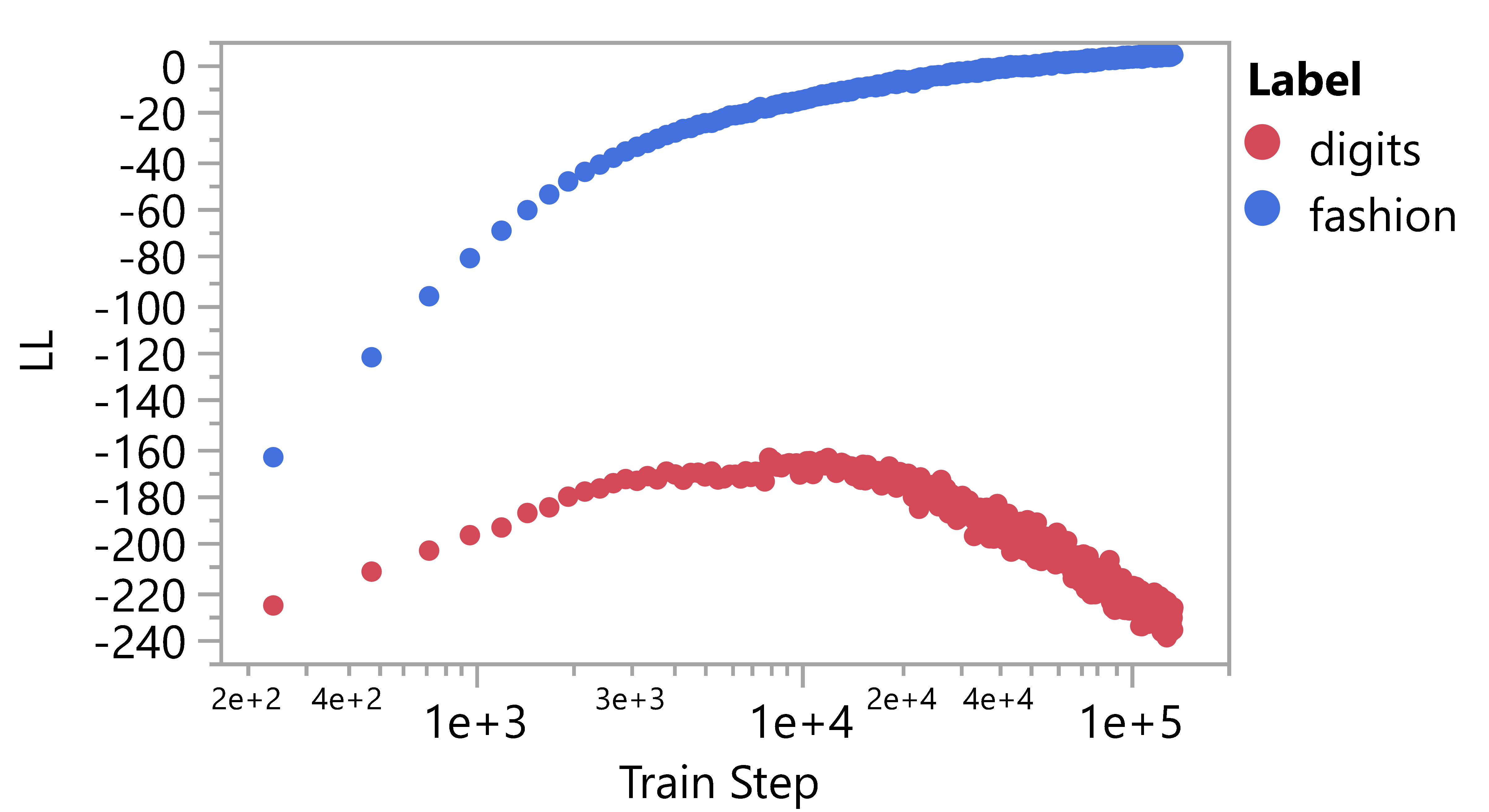}
    \caption{BNAF loss during training.}
    \label{BNAF_training_iter_svd100}
  \end{subfigure}
  \hfill
  \begin{subfigure}[b]{0.42\textwidth}
    \includegraphics[width=\textwidth]{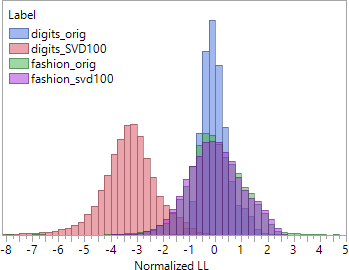}
    \caption{BNAF LL Before/After SVD100}
    \label{fig:BNAF_before_after_SVD_hist}
  \end{subfigure}
  \caption{\label{BNAF_RD_context}	The improvements of the BNAF architecture in modeling the PDF of fashion MNIST are put further in context, showing the expected training process now matching reality (\ref{BNAF_training_iter_svd100}) and the LL scores of the OoD data pushed down considerably (\ref{fig:BNAF_before_after_SVD_hist}).}
\end{figure}

In the process of creating architectures with higher modeling capacity, the previous generation of generative models has all but been forgotten.  The positive results obtained from the full-covariance Gaussian, as well as the improvements from projecting the data to an orthonormal basis, inspires revisiting an old friend.  Probabilistic PCA (PPCA) supports parameterizing a restricted-covariance Gaussian model \citep{bishop:06:pattern} that is more flexible than a diagonal covariance Gaussian and has fewer parameters than a full-covariance Gaussian.  PPCA is very closely related to factor analysis (FA), with the difference primarily in the noise term $\sigma$ as shown in Equation \ref{eq2}.  Where PPCA constrains the error covariance structure to be isotropic, in FA it is relaxed to be an arbitrary diagonal matrix.
\begin{equation} \label{eq2}
\begin{split}
\textbf{z}\sim \mathcal{N}\left( \textbf{\textit{0}}, \textbf{\textit{I}} \right) \\
\textbf{x}|\textbf{z}\sim \mathcal{N}\left(\textbf{W}\textbf{z}+\boldsymbol{\mu}, \sigma^{2}\textbf{I} \right)\\
or\: equivalently\\
\textbf{x}\sim \mathcal{N}\left(\boldsymbol{\mu},\thinspace \sigma^{2}\textbf{I} + \textbf{W}\textbf{W}^{T}  \right)\\
\end{split}
\end{equation}
An expectation maximization (EM) algorithm is used to find \textbf{W} and $\sigma$.  In practice, if anomaly detection and novel data identification are desirable then simply performing PCA and modeling the density from a fraction of the total dimensions seems to work exceptionally well for all models investigated in this research, all while maintaining a fully unsupervised approach (Table \ref{PPCAmnistTable}, Table \ref{PPCAcifarTable}).  PPCA models are especially attractive due to their simplicity, and also train quickly and are available in most statistical software packages.  The results also appear to be very stable/reliable in contrast to the deep generative models, at least on the data-sets tested in this work.
\noindent
\begin{table}[H]
\caption{Reduced-dimensional PDF modeling results for fashion VS digits MNIST (train on fashion) (values in parentheses denote results for models trained on non-quantized values).}
\begin{tabularx}{\textwidth}{@{}lYYYYYY@{}}
\hline
  \textbf{Model} & \# Comp. & \textbf{Test} &  \textbf{OoD}  & Samp & AUC & \#Params\\
\hline
PPCA&100/784&59.4&-614.9&60.5&0.958&5k\\
$MAF_5$&100/784&-55.4 (-55.7)&-164.9 (-165.4)&-55.4 (-57.1)&0.978 (0.98)&153k\\
$MAF_{10}$&100/784&-25.3 (-19.9)&-137.3 (-134.8)&-21.6 (-11.5)&0.978 (0.975)&13.6M\\
BNAF&100/784&4.7 (6.3)&-226 (-204)&N/A&0.991 (0.991)&743k\\
\hline
\end{tabularx}
\label{PPCAmnistTable}
\end{table}
\noindent
\vspace{-4pt}
\begin{table}[H]
\caption{Reduced-dimensional PDF modeling results for CIFAR10 VS SVHN (train on CIFAR10) (values in parentheses denote results for models trained on non-quantized values).}
\begin{tabularx}{\textwidth}{@{}lYYYYYY@{}}
\hline
  \textbf{Model} & \# Comp. & \textbf{Test} &  \textbf{OoD}  & Samp & AUC & \#Params\\
\hline
PPCA&2500/3072&5007&-227.3k&5208&1.0&3.1M\\
$MAF_5$&2500/3072&3442 (3479)&-4280 (-35K)&3526 (3574)&1.0 (1.0)&3.8M\\
$MAF_{10}$&2500/3072&3497 (3498)&-624 (-13k)&3542 (3539)&1.0 (1.0)&87.4M\\
\hline
\end{tabularx}
\label{PPCAcifarTable}
\end{table}
\noindent
Since BNAF in Table \ref{PPCAmnistTable} showed the most significant performance gains in moving from the baseline training to modeling a reduced-dimensional representation with an orthogonal basis, some more details and context are given in Figure \ref{BNAF_RD_context}.  Specifically, we now see the training procedure behaves as expected, with the LL of both data-sets initially increasing and then the OoD LL rapidly falling off as LL is contracted around the data manifold (Figure \ref{BNAF_training_iter_svd100}).  Figure \ref{fig:BNAF_before_after_SVD_hist} also illustrates how before changing basis the distributions of LL scores of in-sample and OoD data were very similar, while after a change of basis the OoD LL was pushed far down with relatively little overlap for the in-sample data.  Appendix \ref{app:granularity} furthermore shows that the trained model ranks the OoD data in intuitive and reasonable ways, further bolstering the methods and techniques presented here for improving PDF modeling with deep generative models.
\section{To Be Continued....}
While likelihood is still a useful (and sometimes only) option to optimize generative models, it remains a poor indicator of sample quality and is furthermore shown in this work to be a poor indicator of OoD data identification for high-dimensional image data.  Diagnostics such as evaluating the sample LL of models and comparing to the training data LL can help identify poor model fits.  Comparing the results of deep generative models to a full-covariance Normal distribution, or a reduced-dimensional PPCA model can also provide helpful safeguards against the poor generalization of deep generative models as anomaly detectors.  Similarly, projecting to an orthonormal basis and/or reducing dimensionality of the data can also hedge against poor generalization for novel data detection.  These results emphasize the need to focus on making architectures and data more amenable to optimization and potentially avoiding models such as PixelCNN++, GLOW, and versions of realNVP that utilize convolutional neural networks for density estimation in high-dimensional images until and if the issues with high likelihood scores of OoD data are resolved for such models.  There may also be advantages in using the methods presented here for unsupervised representation learning in variational autoencoders, and areas such as improving disentanglement of latent factors should be explored.
%\vspace{-15mm}
% Acknowledgements should go at the end, before appendices and references

\acks{We gratefully acknowledge the support of NVIDIA Corporation with the donation of the Titan V GPUs used for this research.” }

\newpage
%\vspace{-5mm}
\appendix
\section{Granularity of Density Estimators}
\label{app:granularity}
\begin{figure}[H]
  \centering
    \begin{subfigure}[b]{0.5\textwidth}
    \includegraphics[width=\textwidth]{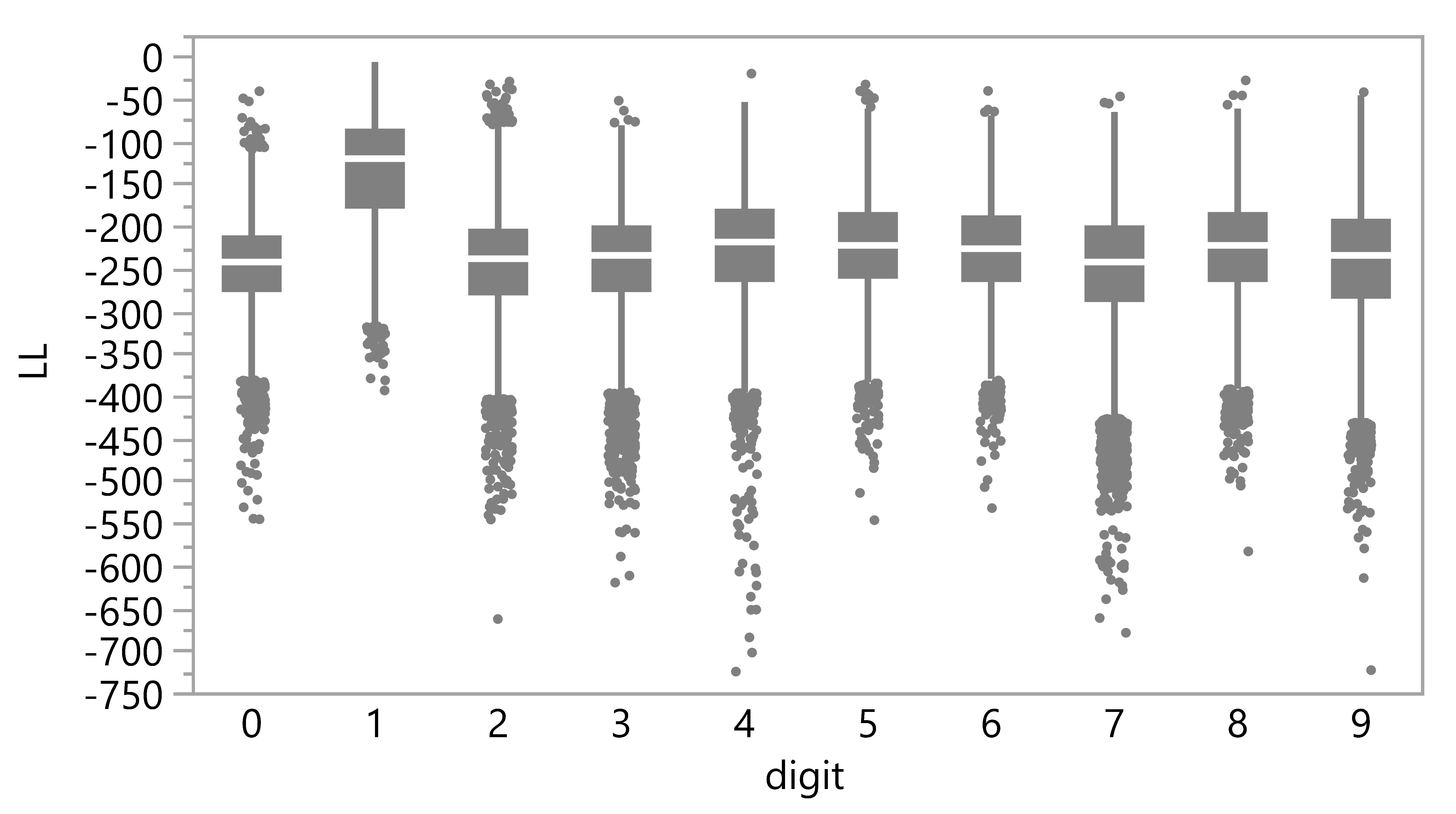}
    \caption{MNIST LL by digit}
    \label{fig:granularity_boxplot}
  \end{subfigure}
  \hfill
  \begin{subfigure}[b]{0.16\textwidth}
    \includegraphics[width=\textwidth]{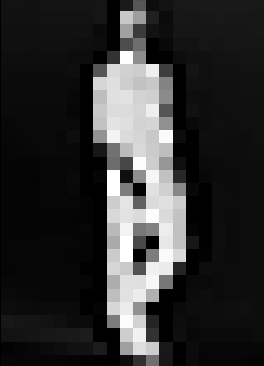}
    \caption{``trousers"}
    \label{fig:trousers}
  \end{subfigure}
  \hfill
  \begin{subfigure}[b]{0.15\textwidth}
    \includegraphics[width=\textwidth]{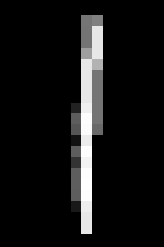}
    \caption{``one"}
    \label{fig:ones}
  \end{subfigure}
  \caption{\label{BNAF_MNIST_granularity}	The LL as evaluated on MNIST digits by the BNAF density model train on fashion MNIST using the first 100 components of the SVD basis vectors is examined for further intuition regarding extrapolation and OoD data.  When the model is properly trained as has been accomplished in this work, the LL assigned to OoD data produces reasonable results, with more visually similar OoD data assigned higher likelihood than other OoD data.}
\end{figure}

The granularity of the LL estimation is one of the least studied properties of generative models since it is a very hard property to judge quantitatively.  However, it is one of the most important properties as an anomaly detector since we usually wish to have control over thresholds that define which data points are anomalies and which are not, as well as having some kind of magnitude regarding how novel a new data point is.  An interesting result was seen by plotting the LL evaluated by class for the MNIST digits data using the BNAF SVD100 model trained with fashion MNIST as shown in Figure \ref{fig:granularity_boxplot}.  All of the digits except for ones had roughly the same LL range, with the ones assigned a significantly higher median likelihood than everything else.  Upon further inspection, this was found to be due to the similarity of the ``ones" class with the ``trousers" class from fashion MNIST, with examples in Figures \ref{fig:trousers}, \ref{fig:ones}.  The trouser example shown in Figure \ref{fig:trousers} was found as one of the mis-labeled images by UMAP (UMAP grouped it with the MNIST ``ones" class).  Thus the extrapolation of the anomaly detector seems to be doing reasonable things in this regard, further bolstering confidence in the methods presented in this work.
% Note: in this sample, the section number is hard-coded in. Following
% proper LaTeX conventions, it should properly be coded as a reference:
%In this appendix we prove the following theorem from
%Section~\ref{sec:textree-generalization}:
\vskip 0.2in
\newpage
\bibliography{references}

\begin{thebibliography}{35}
\providecommand{\natexlab}[1]{#1}
\providecommand{\url}[1]{\texttt{#1}}
\expandafter\ifx\csname urlstyle\endcsname\relax
  \providecommand{\doi}[1]{doi: #1}\else
  \providecommand{\doi}{doi: \begingroup \urlstyle{rm}\Url}\fi

\bibitem[Aapo~Hyvarinen and Oja(2001)]{ICA:01}
Juha~Karhunen Aapo~Hyvarinen and Erkki Oja.
\newblock \emph{Independent Component Analysis}.
\newblock John Wiley and Sons, Inc., New York City, NY, 2001.

\bibitem[Alipanahi et~al.(2015)Alipanahi, Delong, Weirauch, and
  Frey]{alipanahi2015predicting}
Babak Alipanahi, Andrew Delong, Matthew~T Weirauch, and Brendan~J Frey.
\newblock Predicting the sequence specificities of dna-and rna-binding proteins
  by deep learning.
\newblock \emph{Nature biotechnology}, 33\penalty0 (8):\penalty0 831, 2015.

\bibitem[Alireza~Shafaei(2019)]{shafaei:19}
James J.~Little Alireza~Shafaei, Mark~Schmidt.
\newblock A less biased evaluation of out-of-distribution sample detectors, 8
  2019.

\bibitem[Bishop(2006)]{bishop:06:pattern}
Christopher~M. Bishop.
\newblock \emph{Pattern Recognition and Machine Learning}.
\newblock Springer, New York City, NY, 2006.

\bibitem[Bishop(1994)]{bishop:94:mdn}
C.M. Bishop.
\newblock Mixture density networks.
\newblock \emph{Technical Report NCRG 4288, Neural Computing Research Group},
  1994.

\bibitem[Dan~Hendrycks(2019)]{hendrycks:19:anom}
Thomas~Dietterich Dan~Hendrycks, Mantas~Mazeika.
\newblock Deep anomaly detection with outlier exposure.
\newblock ICLR, 2019.

\bibitem[De~Cao et~al.(2019)De~Cao, Titov, and Aziz]{bnaf19}
Nicola De~Cao, Ivan Titov, and Wilker Aziz.
\newblock Block neural autoregressive flow.
\newblock \emph{35th Conference on Uncertainty in Artificial Intelligence
  (UAI19)}, 2019.

\bibitem[Eric~Nalisnick(2019)]{nalisnick:19:anom}
Yee Teh Whye Dilian Gorur Blaji~Lakshminarayanan Eric~Nalisnick,
  Akihiro~Matsukawa.
\newblock Do deep generative models know what they don’t know?
\newblock ICLR, 2019.

\bibitem[Esteva et~al.(2017)Esteva, Kuprel, Novoa, Ko, Swetter, Blau, and
  Thrun]{esteva2017dermatologist}
Andre Esteva, Brett Kuprel, Roberto~A Novoa, Justin Ko, Susan~M Swetter,
  Helen~M Blau, and Sebastian Thrun.
\newblock Dermatologist-level classification of skin cancer with deep neural
  networks.
\newblock \emph{Nature}, 542\penalty0 (7639):\penalty0 115, 2017.

\bibitem[Ghosal(2017)]{ghosal2017engineering}
Sambuddha Ghosal.
\newblock Engineering analytics through explainable deep learning.
\newblock 2017.

\bibitem[Ghosal et~al.()Ghosal, Akintayo, Boor, and Sarkar]{ghosal2017high}
Sambuddha Ghosal, Adedotun Akintayo, Paige Boor, and Soumik Sarkar.
\newblock High speed video-based health monitoring using 3d deep learning.

\bibitem[Ghosal et~al.(2018)Ghosal, Blystone, Singh, Ganapathysubramanian,
  Singh, and Sarkar]{ghosal2018explainable}
Sambuddha Ghosal, David Blystone, Asheesh~K Singh, Baskar Ganapathysubramanian,
  Arti Singh, and Soumik Sarkar.
\newblock An explainable deep machine vision framework for plant stress
  phenotyping.
\newblock \emph{Proceedings of the National Academy of Sciences}, 115\penalty0
  (18):\penalty0 4613--4618, 2018.

\bibitem[Ghosal et~al.(2019)Ghosal, Zheng, Chapman, Potgieter, Jordan, Wang,
  Singh, Singh, Hirafuji, Ninomiya, et~al.]{ghosal2019weakly}
Sambuddha Ghosal, Bangyou Zheng, Scott~C Chapman, Andries~B Potgieter, David~R
  Jordan, Xuemin Wang, Asheesh~K Singh, Arti Singh, Masayuki Hirafuji, Seishi
  Ninomiya, et~al.
\newblock A weakly supervised deep learning framework for sorghum head
  detection and counting.
\newblock \emph{Plant Phenomics}, 2019:\penalty0 1525874, 2019.

\bibitem[Glorot and Bengio(2010)]{glorot:10}
Xavier Glorot and Yoshua Bengio.
\newblock Understanding the difficulty of training deep feedforward neural
  networks.
\newblock pages 9:249--256, 2010.

\bibitem[Hyunsun~Choi(2019)]{choi:19:waic}
A.~Alexander Hyunsun~Choi, Eric~Jan.
\newblock Waic, but why? generative ensembles for robust anomaly detection.
\newblock arXiv:1810.01392, 5 2019.

\bibitem[Ioffe and Szegedy(2015)]{batchnorm:15}
Sergey Ioffe and Christian Szegedy.
\newblock Batch normalization: accelerating deep network training by reducing
  internal covariate shift.
\newblock pages 37:448--456, 2015.

\bibitem[Kingma and Dhariwal(2018)]{glow:2018}
D.~P. Kingma and P.~Dhariwal.
\newblock Glow: Generative flow with invertible 1x1 convolutions.
\newblock page 10236–10245, 2018.

\bibitem[Kingma and Welling(2013)]{vae:13}
D.~P. Kingma and M.~Welling.
\newblock Auto-encoding variational bayes.
\newblock 2013.

\bibitem[Kingma and Ba(2015)]{ADAM15}
Diederik~P. Kingma and Jimmy~Lei Ba.
\newblock Adam: A method for stochastic optimization.
\newblock 2015.

\bibitem[L.~Dinh and Bengio(2014)]{dinh:2014:nice}
D.~Krueger L.~Dinh and Y.~Bengio.
\newblock Nice: Non-linear independent components estimation.
\newblock \emph{ArXiv e-prints}, 2014.

\bibitem[L.~Theis and van~den Oord(2016)]{noteGenEval:16}
M.~Bethge L.~Theis and A.~van~den Oord.
\newblock A note on the evaluation of generative models.
\newblock pages 448--456, 2016.

\bibitem[Laurent~Dinh and Bengio(2017)]{dinh:17:realNVP}
Jascha Sohl-Dickstein Laurent~Dinh and Samy Bengio.
\newblock Density estimation using real nvp.
\newblock 2017.

\bibitem[Liu et~al.(2016{\natexlab{a}})Liu, Ghosal, Jiang, and
  Sarkar]{liu2016unsupervised}
Chao Liu, Sambuddha Ghosal, Zhanhong Jiang, and Soumik Sarkar.
\newblock An unsupervised spatiotemporal graphical modeling approach to anomaly
  detection in distributed cps.
\newblock In \emph{2016 ACM/IEEE 7th International Conference on Cyber-Physical
  Systems (ICCPS)}, pages 1--10. IEEE, 2016{\natexlab{a}}.

\bibitem[Liu et~al.(2016{\natexlab{b}})Liu, Ghosal, Jiang, and
  Sarkar]{liu2016unsupervisedB}
Chao Liu, Sambuddha Ghosal, Zhanhong Jiang, and Soumik Sarkar.
\newblock An unsupervised spatiotemporal graphical modeling approach to anomaly
  detection in distributed cps. in 2016 acm/ieee 7th international conference
  on cyber-physical systems (iccps), pages1--10, 2016{\natexlab{b}}.

\bibitem[Liu et~al.(2017)Liu, Ghosal, Jiang, and Sarkar]{liu2017unsupervised}
Chao Liu, Sambuddha Ghosal, Zhanhong Jiang, and Soumik Sarkar.
\newblock An unsupervised anomaly detection approach using energy-based
  spatiotemporal graphical modeling.
\newblock \emph{Cyber-physical systems}, 3\penalty0 (1-4):\penalty0 66--102,
  2017.

\bibitem[M.~Germain and Larochelle(2015)]{germain:19:made}
I.~Murray M.~Germain, K.~Gregor and H.~Larochelle.
\newblock Made: Masked autoencoder for distribution estimation.
\newblock page 881–889, 2015.

\bibitem[{McInnes} et~al.(2018){McInnes}, {Healy}, and
  {Melville}]{2018arXivUMAP}
L.~{McInnes}, J.~{Healy}, and J.~{Melville}.
\newblock {UMAP: Uniform Manifold Approximation and Projection for Dimension
  Reduction}.
\newblock \emph{ArXiv e-prints}, February 2018.

\bibitem[Mnih et~al.(2015)Mnih, Kavukcuoglu, Silver, Rusu, Veness, Bellemare,
  Graves, Riedmiller, Fidjeland, Ostrovski, et~al.]{mnih2015human}
Volodymyr Mnih, Koray Kavukcuoglu, David Silver, Andrei~A Rusu, Joel Veness,
  Marc~G Bellemare, Alex Graves, Martin Riedmiller, Andreas~K Fidjeland, Georg
  Ostrovski, et~al.
\newblock Human-level control through deep reinforcement learning.
\newblock \emph{Nature}, 518\penalty0 (7540):\penalty0 529, 2015.

\bibitem[Papamakarios et~al.(2017)Papamakarios, Pavlakou, and
  Murray]{papamakarios:2017:maf}
George Papamakarios, Theo Pavlakou, and Iain Murray.
\newblock Masked autoregressive flow for density estimation.
\newblock \emph{Advances in Neural Information Processing Systems}, 2017.

\bibitem[Pokuri et~al.(2019)Pokuri, Ghosal, Kokate, Sarkar, and
  Ganapathysubramanian]{pokuri2019interpretable}
Balaji Sesha~Sarath Pokuri, Sambuddha Ghosal, Apurva Kokate, Soumik Sarkar, and
  Baskar Ganapathysubramanian.
\newblock Interpretable deep learning for guided microstructure-property
  explorations in photovoltaics.
\newblock \emph{npj Computational Materials}, 5\penalty0 (1):\penalty0 1--11,
  2019.

\bibitem[Radford et~al.(2015)Radford, Metz, and
  Chintala]{radford2015unsupervised}
Alec Radford, Luke Metz, and Soumith Chintala.
\newblock Unsupervised representation learning with deep convolutional
  generative adversarial networks.
\newblock \emph{arXiv preprint arXiv:1511.06434}, 2015.

\bibitem[Salimans et~al.(2017)Salimans, Karpathy, Chen, and
  Kingma]{Salimans2017PixeCNN}
Tim Salimans, Andrej Karpathy, Xi~Chen, and Diederik~P. Kingma.
\newblock Pixelcnn++: A pixelcnn implementation with discretized logistic
  mixture likelihood and other modifications.
\newblock In \emph{ICLR}, 2017.

\bibitem[Shah et~al.(2019)Shah, Joshi, Ghosal, Pokuri, Sarkar,
  Ganapathysubramanian, and Hegde]{shah2019encoding}
Viraj Shah, Ameya Joshi, Sambuddha Ghosal, Balaji Pokuri, Soumik Sarkar, Baskar
  Ganapathysubramanian, and Chinmay Hegde.
\newblock Encoding invariances in deep generative models.
\newblock \emph{arXiv preprint arXiv:1906.01626}, 2019.

\bibitem[Silver et~al.(2016)Silver, Huang, Maddison, Guez, Sifre, Van
  Den~Driessche, Schrittwieser, Antonoglou, Panneershelvam, Lanctot,
  et~al.]{silver2016mastering}
David Silver, Aja Huang, Chris~J Maddison, Arthur Guez, Laurent Sifre, George
  Van Den~Driessche, Julian Schrittwieser, Ioannis Antonoglou, Veda
  Panneershelvam, Marc Lanctot, et~al.
\newblock Mastering the game of go with deep neural networks and tree search.
\newblock \emph{nature}, 529\penalty0 (7587):\penalty0 484, 2016.

\bibitem[Yamins and DiCarlo(2016)]{yamins2016using}
Daniel~LK Yamins and James~J DiCarlo.
\newblock Using goal-driven deep learning models to understand sensory cortex.
\newblock \emph{Nature neuroscience}, 19\penalty0 (3):\penalty0 356, 2016.

\end{thebibliography}

\end{document}